\documentclass[runningheads]{llncs}
\usepackage[T1]{fontenc}
\usepackage{graphicx, amsmath, multirow, booktabs, xcolor, float, subcaption}

\begin{document}
\title{Emotion Recognition in Contemporary Dance Performances Using Laban Movement Analysis}
\titlerunning{Emotion Recognition in Contemporary Dance Performances using LMA}
%
\author{Muhammad Turab\inst{1} \and
Philippe Colantoni\inst{1}\orcidID{0000-0003-0002-4435} \and Damien Muselet\inst{1} \orcidID{0000-0001-7803-1171} \and Alain Tr\'{e}meau\inst{1}\orcidID{0000-0003-2826-7519}}
%
\institute{
Laboratoire Hubert Curien - UMR 5516, Saint-Etienne, France \\
\email{muhammad.turab.muslim.bajeer@etu.univ-st-etienne.fr} \\
\email{\{philippe.colantoni, damien.muselet, alain.tremeau\}@univ-st-etienne.fr}
}
\maketitle      

\begin{abstract}
This paper presents a novel framework for emotion recognition in contemporary dance by improving existing Laban Movement Analysis (LMA) feature descriptors and introducing robust, novel descriptors that capture both quantitative and qualitative aspects of the movement. Our approach extracts expressive characteristics from 3D keypoints data of professional dancers performing contemporary dance under various emotional states, and train multiple classifiers, including Random Forests and Support Vector Machines. Additionally, provide in-depth explanation of features and their impact on model predictions using explainable machine learning methods. Overall, our study improves emotion recognition in contemporary dance and offers promising applications in performance analysis, dance training, and human–computer interaction with highest accuracy of 96.85\%.


\end{abstract}
\section{Introduction}
\label{sec:introduction}
In recent years, human emotion recognition has become an important research direction in the field of motion analysis and human-computer interaction. Emotion recognition involves the analysis of various emotional data by extracting features that describe the emotion and their relationship to classify the emotional state. It has been widely used in many fields including emotions in facial expressions \cite{canal2022survey}, speech recognition \cite{khalil2019speech}, text \cite{alswaidan2020survey} and psychological signals \cite{egger2019emotion}. Emotion analysis is not limited to these fields, but it can expand to any field that involves movements. For instance, dance is a form of body movements that is used to express various emotions. Dance has been an integral part of human history, traditions, and cultures to express, tell stories, and convey emotions. Over the years, dance styles have evolved greatly reflecting cultural, social, and artistic changes. One of the new styles that has emerged is contemporary dance that combines elements of jazz, ballet, and modern dance. It originates from a desire to break away from the rigid structures of traditional ballet and modern dance. Unlike traditional dance, it allows dancers to explore novel forms of movement and emotional expression without setting strict rules. Due to the free-form structure of the contemporary dance style, it becomes challenging to recognize emotions. 

For such challenges, Laban Movement analysis (LMA) \cite{von1950mastery} provides a comprehensive framework for the evaluation and interpretation of human body movements. It is divided into four main components including Body, Effort, Shape, and Space. These components describe the structural, geometrical, and dynamic properties of motion. Body and Space components describe how the human body moves, either within the body or in relation with the 3D space surrounding the body. The Shape component describes the shape morphology of the body during the motion, whereas the Effort component focuses on the qualitative aspects of the movement in terms of dynamics, energy, and intent. LMA is more qualitative than quantitative, and a few studies have tried to propose feature descriptors that quantify the qualities of LMA to some extent. However, there are still a few challenges including capturing temporal dynamics of movement, and clear explanation of feature descriptors. Many current feature descriptors for movement analysis may not fully capture the dynamic characteristics without temporal dynamics. To address these challenges, we propose robust feature descriptors with temporal dynamics to capture the dynamic nuances of LMA qualities with their explanation on model predictions and individual performance. The main contributions of this study are as follows:
\begin{itemize}
    \item We improve existing feature descriptors by adding temporal dynamics to accurately capture movement trends and behaviors within short time windows.
    \item We propose new feature descriptors for the four LMA components that capture rich representations of movements.
    \item We perform a comprehensive evaluation on a contemporary dance dataset using Machine Learning (ML) methods, and analyze the interpretability of our descriptors with eXplainable AI methods.
\end{itemize}

\section{Related Work}
\label{sec:related_work} 
Emotion recognition in dance movements has been an active field due to its applications in human-computer interaction \cite{brave2007emotion}, performance analysis \cite{sun2023deep}, and evaluation \cite{aristidou2015folk}. Understanding how dancers express emotions through movement provides valuable insights into the expressive behavior of the performer. To extract expressive behavior from movement, LMA has become a popular framework as it provides a structured language to describe both the quantitative and qualitative aspects of human motion. In study \cite{ajili2019expressive}, LMA is used to recognize and analyze expressive gestures, including dancing, moving, waving, pointing, and stopping, with four emotions happy, angry, sad and neutral. Feature evaluation was performed using ML methods and subjective evaluation. The results show that the features are 87. 03\% accurate in gesture classification. \cite{tan2025research} used Kinect to track dancers and developed models for action and emotion recognition with 88.34\% and 98.95\% accuracy on skeleton data. Features like eigenvalue speed and skeleton pair distance help distinguish emotions. LMA was used in human action recognition in \cite{ramezanpanah2020human}, with a two-stage approach in which the first feature descriptors are defined using three components of LMA, including body, space, and shape. The results were obtained on various datasets using Dynamic Time Warping (DTW) to measure similarity in actions. Several studies have focused on the classification of dance emotions using LMA-based feature descriptors. A method is proposed to extract the motion qualities from dance performances for indexing and analysis in \cite{aristidou2014feature}. LMA-based features were used to investigate the correlation of features with dancers expressed emotional state.  The results show low inter-feature correlation between dancers and their emotional state. \cite{aristidou2015emotion} investigated similarities of various emotional states using extracted LMA extracted features. The results show that the features can partially extract LMA components that can be used for the dance classification from emotion.

For automatic feature extraction and better results, Deep learning (DL) has been used. \cite{wang2020dance} applied a hybrid deep learning method based on Convolutional Neural Networks (CNN) and Long Short-Term Memory (LSTM) to effectively identify emotions. For body structure, spatial orientation, and force effect LMA-based feature descriptors were used to add emotional changes during movement. Their approach successfully recognizes emotions in dance movements with high accuracy.

\section{Methodology} 
\label{sec:methodology}

\begin{figure}
    \centering
    \includegraphics[width=\textwidth]{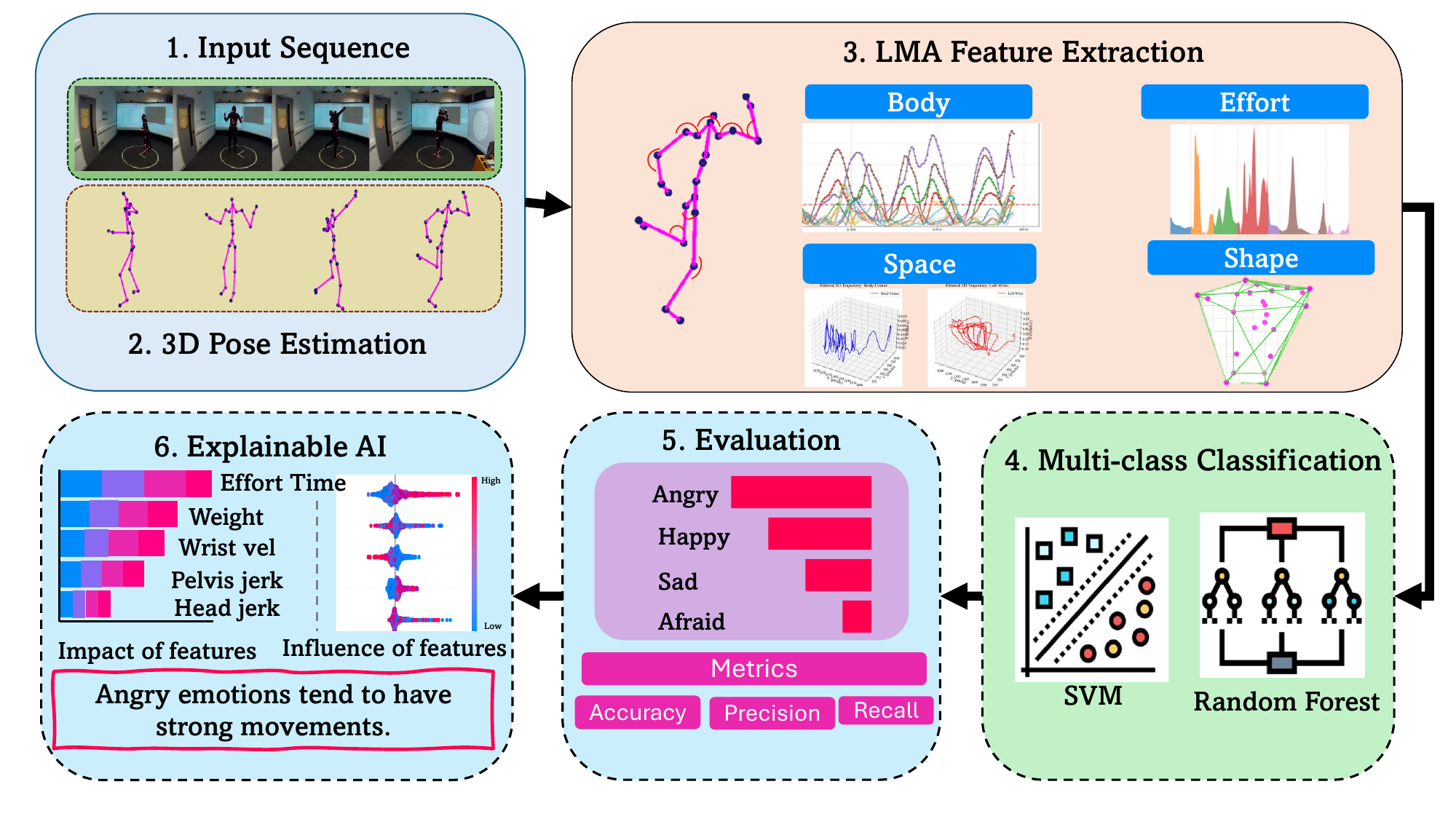}
    \caption{Overview of the proposed method for Emotion Recognition in Contemporary Dance Performances.}
    \label{fig:workflow}
\end{figure}

To improve emotion recognition in contemporary dance, our study is divided into six main parts: 1) Data pre-processing, 2) 3D human body pose estimation, 3) LMA feature extraction, 4) Multiclass classification, 5) Evaluation, and 6) Explainability of feature descriptors as shown in Fig. \ref{fig:workflow}.

\subsection{Data Pre-processing}
\label{subsec:data_processing}
We use contemporary dance performance videos from the Dance Motion capture dataset provided by the University of Cyprus \cite{Aristidou:2019:JOCCH}. This dataset contains performances of 5 dancers expressing 12 emotions including afraid, angry, annoyed, bored, excited, happy, miserable, pleased, relaxed, sad, satisfied and tired. The dataset has a few limitations, such as: 1) inconsistent video duration, 2) limited number of dancers and performances, 3) low frame rate, and 4) a fixed camera angle for a single perspective.


\subsection{3D Human Body Pose Estimation}
To obtain the joints keypoint data from dance videos, we use Neural Localizer Fields (NLF) \cite{sarandi2024neural} for 3D body pose estimation. We compare NLF with other state-of-the-art (SOTA) methods including Mediapipe \cite{lugaresi2019mediapipe} and OpenPose \cite{cao2019openpose} and choose NLF for several reasons including: 1) It is an advanced model trained on massive data to estimate 2D/3D points from single image, 2) the dance performances have many frames where some body parts are occluded or missing in some cases when dancer is very close to the camera and NLF performs really well in such cases outperforming existing SOTA methods. 3) Dancers back-and-forth movement introduces challenges in feature extraction due to shifting camera perspectives. NLF effectively addresses shifting camera perspective by providing the 3D keypoints in absolute camera-space.


\subsection{LMA Feature Extraction}
Laban Movement Analysis is a language and a structured framework to describe, and interpret human body movements. It categorizes human motion into four main components: BODY, EFFORT, SHAPE and SPACE. \textbf{Body} component describes the structural and physical characteristics of human body and is responsible for joints connectivity, movement and their influence on each other. For joint connectivity, we calculate the Euclidean distance and angles between the hands, shoulders, pelvis, knees, and ankles. For movement, we propose a new feature descriptor to describe the joint that initiates movement in a sequence, as shown in Equation \ref{eq:movement_initiation}.
\begin{equation}
\text{Initiation}(t) = \frac{P_j\bigl(t + w\bigr) - P_j\bigl(t\bigr)}{\Delta t} > \tau
\label{eq:movement_initiation}
\end{equation}



Here \(P_{j}(t)\) denotes the position of joint \(j\) at time \(t\), \(\Delta t\) is the time interval, \(w\) is the short time-window to add temporal dynamics, and \(\tau\) is a data-driven threshold calculated using standard-deviation of the entire sequence. \textbf{Effort} component describes the intention, dynamic qualities of movement, and energy used during movement. It is associated with the change in emotion or mood, hence it is useful for motion expressivity and emotion description. We mainly focus on upper and lower body parts, such as hands, feet, head, and pelvis, as these joints are involved most in emotion expression. The Effort component has four factors: Space, Weight, Time, and Flow. Effort Space describes the attention of movement in space, and it can be direct where the movement is focused on single direction, and indirect where the movement is focused in multi-directions. For this we calculate the distance of each joint for a short time-window to the total distance covered by that joint using Equation \ref{eq:space_joint}.
\begin{equation}
\text{Space}_j(T) = \frac{\sum_{i=1}^{T} \left\| P_j(t_i) - P_j(t_i - w) \right\|}{\left\| P_j(T) - P_j(t_1) \right\|}
\label{eq:space_joint}
\end{equation}
Total space is calculated by multiplying the weights \(\alpha_j\)  of selected joints \(\mathcal{J}\) to their space factor using Equation \ref{eq:space_total}. Joint weights are used to give more importance to extremities, as defined in \cite{mmpose2020}.
\begin{equation}
\text{Space}(T) = \sum_{j \in \mathcal{J}} \alpha_j \, \text{Space}_j(T)
\label{eq:space_total}
\end{equation}

Effort Weight describes how powerful or strong a movement is, it can be strong or light. For this we calculate the kinetic energy of selected joints \(\mathcal{J}\) using Equation \ref{eq:effort_weight}. Here \(v_j(t_i)\) denotes the velocity of a joint $j$ at time $t_i$.

\begin{equation}
Weight(t) 
= \sum_{j \in \mathcal{J}} E_j(t_i) 
= \sum_{j \in \mathcal{J}} \frac{1}{2}\alpha_j v_j(t_i)^2
\label{eq:effort_weight}
\end{equation}

Effort Time captures a sense of urgency. It tells how quick or sustained a movement is executed. For this we calculate the acceleration of selected joints over a sliding time-window as shown in Equation \ref{eq:effort_time}.

\begin{equation}
\text{Time}(T) = \sum_{j \in \mathcal{J}} \alpha_j \, \text{Time}_j(T)
\label{eq:effort_time}
\end{equation}
where \(\text{Time}_j(T) = \frac{1}{T} \sum_{i=1}^{T} a_j(t_i)\), and $a_j$ denotes the acceleration of a joint $j$. All these features are inspired from \cite{larboulette2015review}. \textbf{Space} component describes the relationship of movement with space. For this we calculate trajectory of whole sequence, curvature, and propose a new feature spatial dispersion that describes how dancer is using kinesphere or it's personal space. It is calculated as distance of upper body parts to torso, pelvis for lower body parts. total path covered, and total distance covered. \textbf{Shape} component describes how the body is changing shape during the movement. For this we calculate the volume of the body using ConvexHull algorithm implemented in Python SciPy package \cite{virtanen2020scipy}. 

Once we quantify all LMA components, we obtain a descriptor vector composed of 54 features which is used as input to the classification models.

\subsection{Multi-Class Emotion Classification}
For emotion classification we use ML methods including Support Vector Machines (SVM) \cite{cortes1995support}, and Random Forest (RF) \cite{breiman2001random}, since they are already used for emotion classification in \cite{ajili2019expressive}\cite{wang2020dance}\cite{aristidou2015emotion}. Firstly, we use 3 fold cross-validation to divide the dataset into 3 sets including training, testing, and validation to overcome overfitting and data dependency. For both SVM and RF the best hyperparameter values were found using GridSearch \cite{bergstra2012random}, and the best parameter values were selected based on the validation accuracy.

\subsection{Model Explainability Using SHAP}
Machine learning models are often considered black-box systems, which means that the process by which they arrive at specific predictions remains unclear. It is challenging to determine the factors that influence their results. In order to improve our understanding for emotion classification in contemporary dance performances, we use SHapley additive explanations (SHAP) \cite{NIPS2017_7062}, which is a game theory-based method to explain the predictions of ML models by assigning importance values to each feature for a given prediction. It tells the contribution of each feature to a specific prediction. For SVM we applied KernelExplainer, and for Random Forest we applied TreeExplainer \cite{lundberg2020local2global}.

\section{Results}
In this section, we present the experimental results obtained using proposed LMA feature descriptors. For evaluation accuracy, precision, and recall metrics are used to assess the performance of our method. Next, we summarize the key quantitative results, highlighting how our approach compares with other methods. Table \ref{tab:all_emotions} shows the results of individual emotional states, It can be observed that RF demonstrates consistent performance across all classes and metrics. Similarly, the SVM shows balanced performance but is lower than that of RF. Our method maintains consistent performance across all emotional states, unlike \cite{aristidou2015emotion}, where some classes show high accuracy while others perform lower, clearly indicating that the features do not capture overlapping emotions.

\begin{table}
\centering
\setlength{\tabcolsep}{2pt}
\caption{Classification results of RF and SVM for each emotion using sliding window of 25 consecutive frames.}
\label{tab:classification_report}
\begin{tabular}{l@{\hskip 0.5cm}ccc@{\hskip 0.5cm}ccc}
\toprule
\multirow{2}{*}{\textbf{Emotion}} 
 & \multicolumn{3}{c}{\textbf{RF}} 
 & \multicolumn{3}{c}{\textbf{SVM}} \\
\cmidrule(lr){2-4} \cmidrule(lr){5-7}
 & Precision (\%) & Recall (\%) & F1-score (\%) & Precision (\%) & Recall (\%) & F1-score (\%)\\
\midrule
Afraid   & 98.46 & 100.00 & 99.22 & 92.57 & 97.27 & 94.86 \\
Angry    & 94.71 & 96.59 & 95.64 & 91.34 & 92.90 & 92.11 \\
Annoyed  & 98.95 & 93.56 & 96.18 & 95.38 & 92.08 & 93.70 \\
Bored  & 97.62 & 98.80 & 98.20 & 94.74 & 93.98 & 94.35 \\
Excited  & 91.71 & 97.07 & 94.32 & 92.57 & 92.82 & 92.70 \\
Happy  & 97.18 & 92.81 & 94.95 & 93.43 & 93.71 & 93.57\\
Miserable  & 98.27 & 98.61 & 98.44 & 92.96 & 91.67 & 92.31\\
Pleased  & 97.04 & 92.58 & 94.76 & 96.27 & 91.17 & 93.65\\
Relaxed  & 95.81 & 99.38 & 97.56 & 95.99 & 96.58 & 96.28\\
Sad  & 97.71 & 97.99 & 97.85 & 91.69 & 94.84 & 93.24\\
Satisfied  & 99.07 & 99.07 & 99.07 & 97.60 & 94.86 & 96.21 \\
Tired  & 99.57 & 99.13 & 99.35 & 95.28 & 96.10 & 95.69\\
\midrule
\textbf{Average} & \textbf{97.39} & \textbf{96.21} & \textbf{96.73} & \textbf{94.24} & \textbf{93.77} & \textbf{93.98} \\
\bottomrule
\end{tabular}
\label{tab:all_emotions}
\end{table}

Table \ref{tab:comparison} presents the comparison of different methods on the Dance Motion Capture dataset when compared to the proposed method. It is evident that the proposed method outperforms all other methods in terms of accuracy. RF achieves a higher accuracy of 96.85\% compared to SVM's 93.90\%.

\begin{table}
\centering
\caption{Comparison of emotion classification with state-of-the-art ML methods on Dance Motion Dataset \cite{Aristidou:2019:JOCCH} using sliding window of 25.}
\label{tab:comparison}
\setlength{\tabcolsep}{8pt} 
\begin{tabular}{lcc}
\toprule
\textbf{Study}       & \textbf{Method}  & \textbf{Accuracy (\%)} \\
\midrule
\cite{wang2020dance} & Decision Tree    & 94.48 \\
\cite{wang2020dance} & Random Forest    & 92.95 \\
\cite{aristidou2015emotion} & SVM & 89.96 \\
\cite{aristidou2015emotion} & Random Forest & 91.75 \\
\cite{bai2021emodescriptor} & SVM & 81.70 \\
Ours & SVM & 93.90 \\
\textbf{Ours} & Random Forest & \textbf{96.85} \\
\bottomrule
\end{tabular}
\label{tab:comparison}
\end{table}

\begin{figure}
    \centering
    \includegraphics[width=1.1\textwidth, trim=50 220 50 110, clip]
{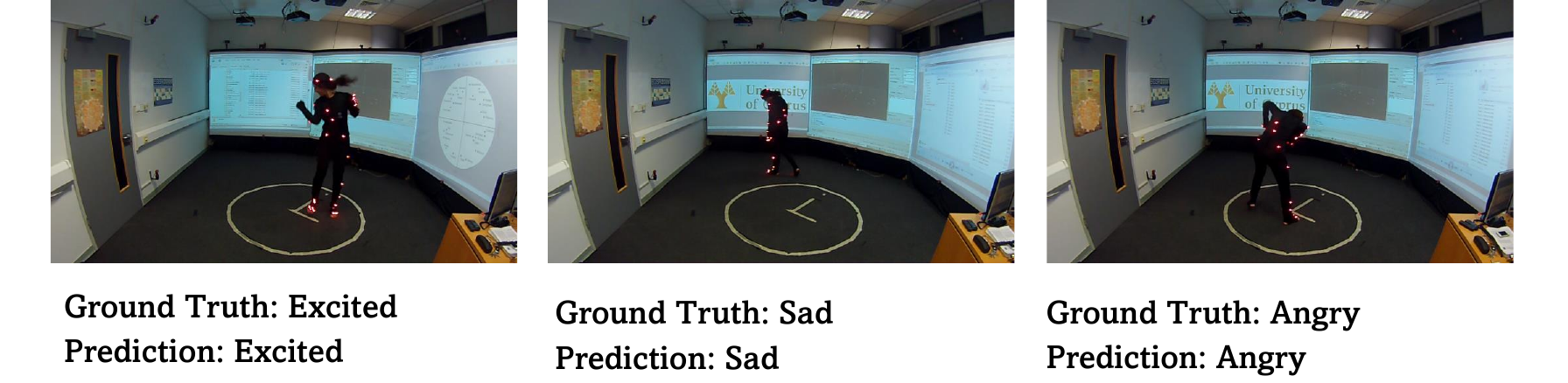}
    \caption{Ground truth and model predictions for three unseen dance performances. Images from \cite{Aristidou:2019:JOCCH}.}
    \label{fig:model_classification}
\end{figure}

In Fig. \ref{fig:model_classification}, we show the predictions of our method for unseen movement sequences, where it successfully recognizes emotions in dance performances.

\begin{figure}
    \centering
    \includegraphics[width=0.57\linewidth]{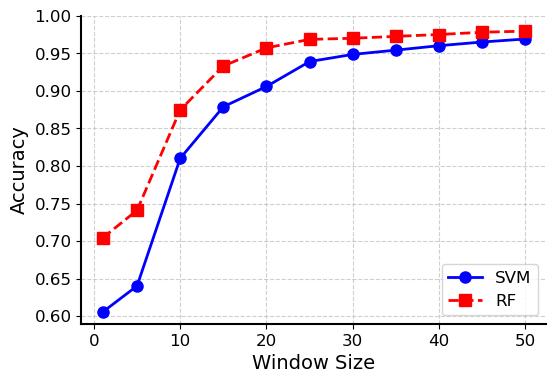}
    \caption{Impact of sliding window size on accuracy.}
    \label{fig:sliding_window}
\end{figure}

Our sliding window approach improves the overall accuracy of emotion classification, as shown in Fig. \ref{fig:sliding_window}. It can be observed that a smaller window size results in lower accuracy, while increasing the window size improves the accuracy for both RF and SVM. In our case, the optimal window size is 25-30 for a balanced performance, and after 30 the improvement is negligible. Although it is evident that a larger window improves accuracy, it is also important to note that it may suppress nuanced emotional details. Therefore, selecting an optimal window size is crucial.

\begin{figure}
    \centering
    \includegraphics[width=0.7\textwidth]{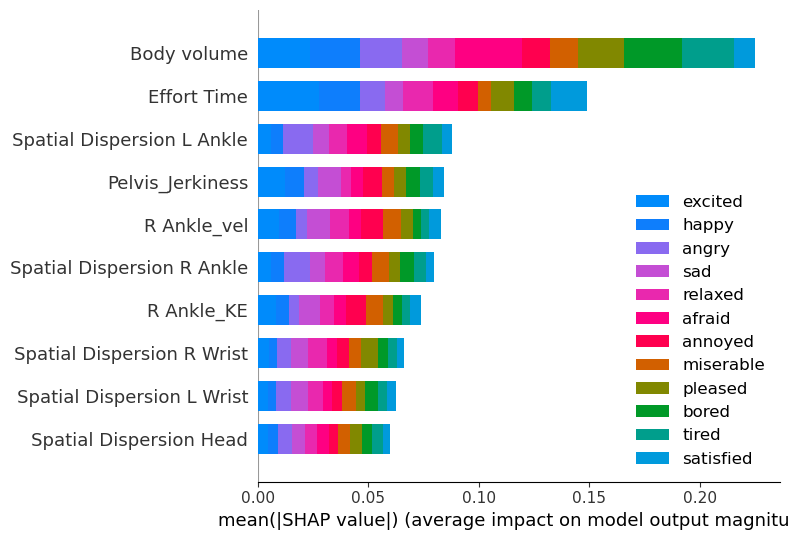}
    \caption{Impact of top 10 features on models predictions.}
    \label{fig:impact}
\end{figure}

Currently, our method achieves promising results for emotion classification in contemporary dance. However, there remains a challenge regarding interpretability and explainability of the models, which has not been explored in existing research on emotion classification. We aim to investigate questions such as which features significantly affect the ability of models to classify angry movements from those that are happy or excited, as they look very similar visually. To answer such questions, we apply various method that explain the contributions of features individually or as a whole. The SHAP summary plot in Fig. \ref{fig:impact} shows most influential features in predicting emotions from movement data. Body volume has the highest impact, indicating that overall body expansion or contraction plays a key role in emotional expression. LMA Effort time has high impact as well which is essential to distinguish angry movements from sad. Overall, both global body features and localized joint behavior impact model's predictions.

\begin{figure}
    \centering
        \centering
        \includegraphics[width=0.74\textwidth]{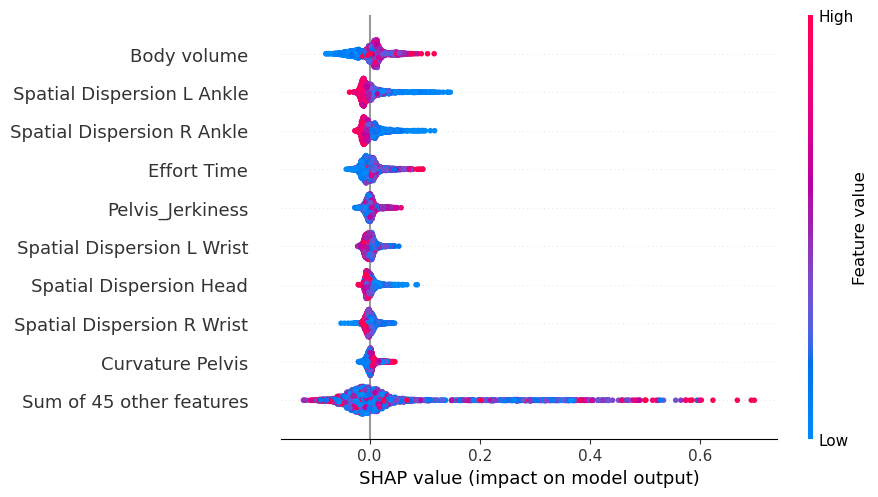}
        \caption{Impact of features on the model's predictions for dance movements expressing anger.}
        \label{fig:individual_influence_angry}
\end{figure}

\begin{figure}
        \centering
        \includegraphics[width=0.74\textwidth]{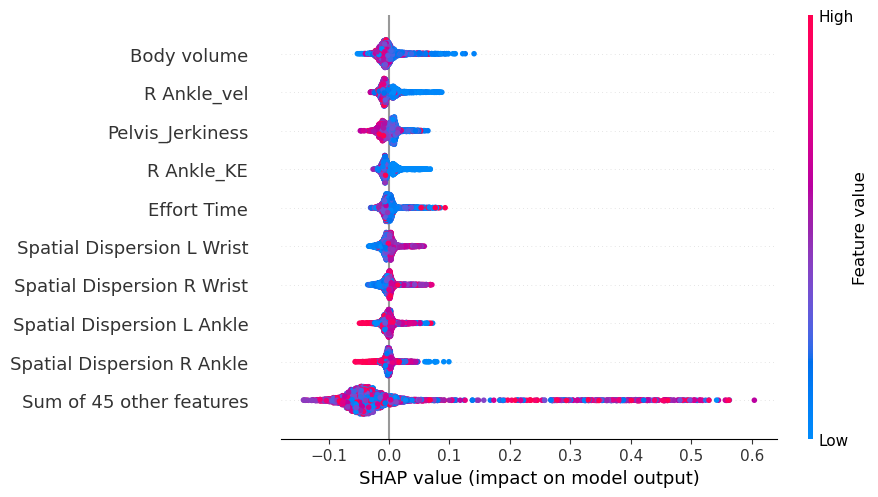}
        \caption{Impact of features on the model's predictions for dance movements expressing sadness.}
    \label{fig:individual_influence_sad}
\end{figure}

Impact of key movement features on emotion predictions are illustrated in Fig.~\ref{fig:individual_influence_angry} and Fig.~\ref{fig:individual_influence_sad}. The horizontal axis represents the SHAP value, where positive values push the prediction towards a target class and negative values push it away. In both plots, features such as \textit{Body volume}, \textit{Effort Time}, and \textit{Pelvis Jerkiness} show strong positive influence, indicating that emotions associated with expansive or intense movements (e.g., anger or excitement) are driven by higher body expansion, energetic motion, and movement irregularities. Conversely, lower values in these features tend to align with sustained movements (e.g., sadness or tiredness), reflecting more contracted posture, reduced movement dispersion, and lower kinetic energy.

\section{Conclusion}
In this study, we improve existing Laban Movement Analysis (LMA) feature descriptors by adding the temporal dynamics with sliding window approach, introduce novel robust descriptors that capture both quantitative and qualitative aspects of emotions in dance. With a comparative analysis, we show that the proposed approach achieves higher accuracy compared to existing methods. Furthermore, we provide interpretations and explanations of the feature descriptors to better understand black-box models and their predictions. Overall, our study improves the understanding and recognition of dance emotions using improved LMA based feature descriptors, and contribute to more effective and transparent emotion and motion analysis in dance performances. Future work includes extending the framework to dance style classification.

\subsubsection{Acknowledgments}
This study was carried out as part of the PREMIERE project "Performing Arts in a new Era", https://premiere-project.eu/, funded by HORIZON-CL2-2021-HERITAGE-000201-04 (grant number 101061303 - PREMIERE). 

%
%
%
%
\bibliographystyle{splncs04}
\bibliography{main}
\end{document}